\documentclass{bmvc2k}
\usepackage[normalem]{ulem}
\usepackage{diagbox}
\usepackage{gensymb}
\usepackage{bm}
\usepackage{amsmath}
\usepackage{pdfpages}

\title{Domain Adaptation of Learned Features \\ for Visual Localization}

\addauthor{Sungyong Baik\textsuperscript{$\dagger$}}{dsybaik@snu.ac.kr}{2}
\addauthor{Hyo Jin Kim}{kimhyojin@fb.com}{1}
\addauthor{Tianwei Shen}{tianweishen@fb.com}{1}
\addauthor{Eddy Ilg}{eddyilg@fb.com}{1}
\addauthor{Kyoung Mu Lee}{kyoungmu@snu.ac.kr}{2}
\addauthor{Chris Sweeney}{sweeneychris@fb.com}{1}
\addinstitution{
 Facebook Reality Labs \\ Redmond, WA, USA
}
\addinstitution{
ASRI, Department of ECE \\
Seoul National University \\
Seoul, South Korea
}

\runninghead{Baik et al.}{Domain Adaptation of Learned Features for Vis. Localization}

\def\etal{\emph{et al}\bmvaOneDot}
\begin{document}

\maketitle

\begin{abstract}
We tackle the problem of visual localization under changing conditions, such as time of day, weather, and seasons.
Recent learned local features based on deep neural networks have shown superior performance over classical hand-crafted local features. However, in a real-world scenario, there often exists a large domain gap between training and target images, which can significantly degrade the localization accuracy.
While existing methods utilize a large amount of data to tackle the problem, we present a novel and practical approach, where only a few examples are needed to reduce the domain gap.
In particular, we propose a few-shot domain adaptation framework for learned local features that deals with varying conditions in visual localization.
The experimental results demonstrate the superior performance over baselines, while using a scarce number of training examples from the target domain. 
\end{abstract}

\section{Introduction}
Visual localization is the problem of estimating the 6-DoF camera pose of a query image, given a set of reference images~\cite{6dof2018sattler}. It is one of the key technologies for robot navigation~\cite{lim2012real,cummins2008fab,li2019cross}, autonomous driving \cite{mcmanus2014shady}, and augmented and virtual reality \cite{middelberg2014scalable,lynen2015get}. 
Visual localization is challenging because often the query and its corresponding reference images are taken in different viewpoints, illumination, time of day, weather, and seasons \cite{milford2012seqslam,linegar2016made,naseer2017semantics}. In addition, occlusions, ubiquitous objects (geographically prevalent objects, such as traffic lights), and transient objects (objects that move or change its appearance, such as cars or trees) can cause confusion in the localization process, making the problem even more difficult~\cite{knopp2010avoiding,kim2017learned,kim2015predicting}.

Recent learned local features \cite{luo2018geodesc,luo2019contextdesc,dusmanu2019d2,tian2019sosnet} based on deep neural networks have demonstrated increased robustness against photometric and geometric changes over traditional local features \cite{sift1999lowe,matas2004robust,bay2006surf}, owing to the higher representative power and the end-to-end training paradigm.
Still, there exists a severe performance gap when the target image domain is different from the training image domain \cite{finegrained2019larsson,6dof2018sattler}. For example, feature extraction models trained on sunny or daytime images do not perform well on snowy or nighttime images. 

To tackle this problem, there has been a recent push on using image translation techniques to convert a source domain image to a target domain image~\cite{nighttoday2019anoosheh,adversarial2018porav}. However, such approaches require a significant amount of training images. Collecting data for all possible conditions is difficult and cumbersome. 
Furthermore, deploying these methods requires an additional routing stage to apply the relevant translation model, which may introduce unrecoverable errors due to mis-classification~\cite{ahmed2016network,kim2018hierarchy,murthy2016deep,warde2014self,yan2015hd}. 

In this work, we instead focus on a more common, practical scenario where only a few examples of target domain images are available, which is often the case in real-world scenarios. 
To this end, we propose a few-shot domain adaptation framework for learned local features. 
To the best of our knowledge, ours is the first work to explore few-shot domain adaptation of CNN-based local features in the context of visual localization. 
Our framework assumes that the reference representation stays fixed, and does not require re-extraction of features for the reference 3D models as a result of domain adaptation, which is computationally heavy. Together with the fact that the training match pairs are automatically generated based on the current representation, it opens up the possibility for continuous online training.
\begin{figure*}[t]
\centering
\includegraphics[width=0.99\textwidth]{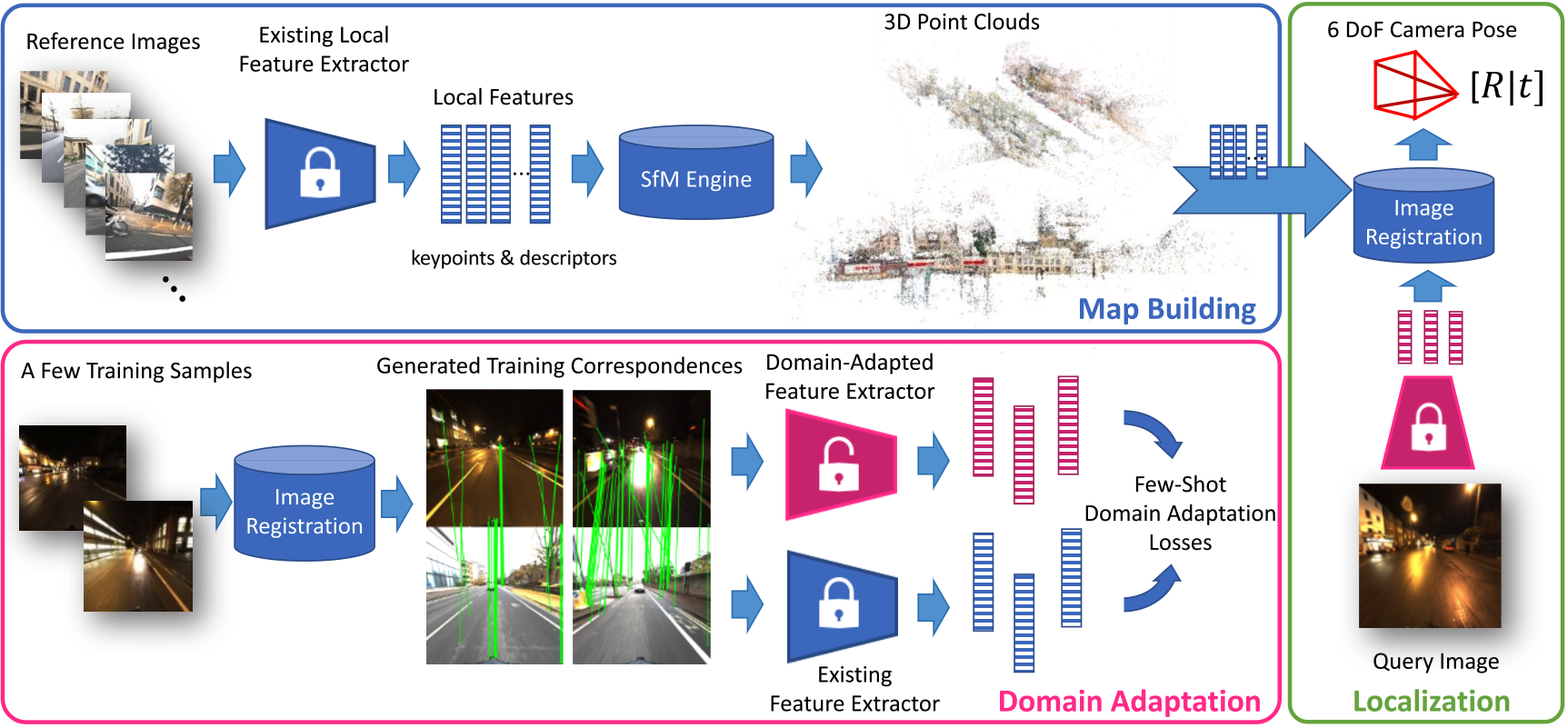}
\vspace{-0.8em}
\caption{Overview of the proposed framework. See Sec \ref{sec:overview} for a detailed description. The lock symbol indicates whether the network parameters are fixed or trained. 
}
\vspace{-0.4cm}
\label{fig:overview}
\end{figure*}
We introduce three objective functions for this task, that can be used in addition to the correspondence loss between the target and the source domain descriptors. 
The first loss aligns the overall distribution of local descriptors from the source and target domains that belong to the same visual element.
Our second loss provides more granular level of constraint by enforcing the pairwise distances between the examples in the source manifold subspace to be similar to that of the corresponding examples in the target subspace.
Finally, we propose an auxiliary loss that penalizes the feature matches that conflict with the given pose. 

We performed evaluation with varying number of training samples on the standard benchmark datasets for visual localization. 
The results demonstrate superior performance of our method over the baselines when only a few examples from the target domain are available. We also provide the comparison of different combinations of the loss terms. 
Beyond the detailed innovations, our framework can be applied to any CNN-based local feature extractor. 

\vspace{-1em}
\section{Related Work}
\textbf{Visual Localization. } 
Traditionally, visual localization is performed by registering the given query image to the 3D point cloud model reconstructed from a set of reference images through the Structure-from-Motion technique \cite{ullman1979interpretation}. 
Specifically, 2D-3D matching is performed based on the local descriptors from the query image and the point cloud. Then, the camera pose is estimated from 2D-3D matches using a PnP solver \cite{haralick1994review}. Efficient matching can be done by prioritized matching \cite{li2010location,sattler2012improving,sattler2015hyperpoints} or by performing image retrieval to reduce the search space \cite{6dof2018sattler,sattler2017large,taira2018inloc,sattler2016large}. This line of approach is often called the 3D-structure-based approach \cite{sattler2017large} and our method also belongs to this category. There are other streams of work such as the 2D-based approach, where the query pose is approximated using the visually similar reference images \cite{arandjelovic2016netvlad,arandjelovic2014dislocation,kim2017learned,knopp2010avoiding,chen2011city}, and the regression-based methods that estimate the camera pose directly from the input image using deep neural networks \cite{brachmann2017dsac,brachmann2018learning,walch2017image,brahmbhatt2018geometry,kendall2015posenet,kendall2017geometric}. 

\noindent
\textbf{Learned Local Features. }
Traditional local features like SIFT~\cite{sift1999lowe} and SURF~\cite{bay2006surf} have dominated the area for almost two decades, due to their generalization ability. Yet, these hand-crafted local features do not utilize the full representation power inherited in the data. There is more and more evidence showing that CNN-based learned local descriptors outperform traditional ones in a number of tasks~\cite{luo2018geodesc,luo2019contextdesc,dusmanu2019d2} by a large margin. 
CNN-based local descriptor learning is typically formulated as a metric learning problem~\cite{Han_2015_CVPR,Zagoruyko_2015_CVPR,Simo-Serra_2015_ICCV}. Different aspects of the problem, such as hard negative mining~\cite{mishchuk2017working}, adding regularization terms~\cite{zhang2017learning,tian2019sosnet}, and incorporating context information~\cite{luo2019contextdesc} were also explored.
There is also a recent trend to jointly optimize the detector and the descriptor~\cite{yi2016lift,detone2018superpoint,dusmanu2019d2}. SuperPoint~\cite{detone2018superpoint} computes both keypoints and descriptors based on the notion of homographic adaptation. D2-Net~\cite{dusmanu2019d2} extends this idea by sharing and optimizing the parameters jointly for the detector and the descriptor.  
The above methods, though addressing learning local features in different perspectives, have fundamental drawbacks in domain generalization ability. 
In this paper, we address the challenge of reducing this domain gap.

\noindent
\textbf{Domain Adaptation for Deep Neural Networks. }
Domain adaptation methods typically aim to find a transformation that aligns the source and the target feature spaces~\cite{fernando2013unsupervised,sun2016deep}.  
There is a large body of literature for shallow models \cite{daume2009frustratingly,tommasi2013frustratingly,khosla2012undoing,hoffman2012discovering,kulis2011you,gopalan2011domain,yang2007adapting,saenko2010adapting,gong2012geodesic}. For deep neural networks, it is traditionally achieved through fine-tuning~\cite{girshick2014rich}. However, fine-tuning often results in over-fitting due to the imbalance between the source and the target domain examples~\cite{tzeng2014deep} and is not applicable when no target label is provided.
To overcome this problem, several methods were proposed~\cite{tzeng2014deep,tzeng2017adversarial,motiian2017few,saito2019semi,ma2019gcan}. These include the unsupervised confusion losses based on the first- and the second-order statistics~\cite{quinonero2008covariate,tzeng2014deep,long2015learning,sun2016deep}, and adversarial losses by training domain classifiers~\cite{tzeng2017adversarial,ghifary2016deep,tzeng2015simultaneous,ganin2014unsupervised,shu2018dirt,motiian2017few}. 

\noindent
\textbf{Domain Adaptation for Visual Localization. }
Our work is mostly related to recent work that focuses on reducing the domain gap between the training and the test images \cite{adversarial2018porav,nighttoday2019anoosheh,imagetoimage2019mueller}. Porav~\etal~\cite{adversarial2018porav}, use CycleGAN \cite{zhu2017unpaired} to transform a source domain image to a target domain image (e.g., night-to-day, winter-to-summer), to enhance the local feature matching. 
In a similar spirit, Anoosheh~\etal introduced ToDayGAN~\cite{nighttoday2019anoosheh} to convert the images from nighttime to daytime to improve the image retrieval stage with DenseVLAD~\cite{torii201524}. 
Mueller~\etal~\cite{imagetoimage2019mueller} use view synthesis to enhance data augmentation. However, all these methods require a significant number of training images and can moreover introduce undesired artifacts. 
Some methods utilize the modalities that are invariant to changes, such as semantic or depth information \cite{toft2018semantic,schonberger2018semantic,mousaviansemantically,naseer2017semantics,shi2019visual}. 
However, getting such modalities consistently from RGB inputs under changing conditions is challenging by itself unless such variances are reflected in the training data \cite{finegrained2019larsson}. 

In this work, we propose a few-shot domain adaptation framework that presents a practical yet effective setting, where only few examples from the target domain are available.
In this new setting, we adopt one of the classical domain adaptation losses, but use pseudo labels (coarse visual words) to further capture the structure of the data, based on the feature correspondences that we generate (Sec. \ref{subsec:vwcoral}). 
We also introduce two additional loss terms (Sec. \ref{subsec:cdsos}-\ref{subsec:softmatch}) that are effective for few-shot domain adaptation of learned local features. 

\vspace{-1em}
\section{Proposed Method}
\begin{figure*}[t]
\centering
\includegraphics[width=\textwidth]{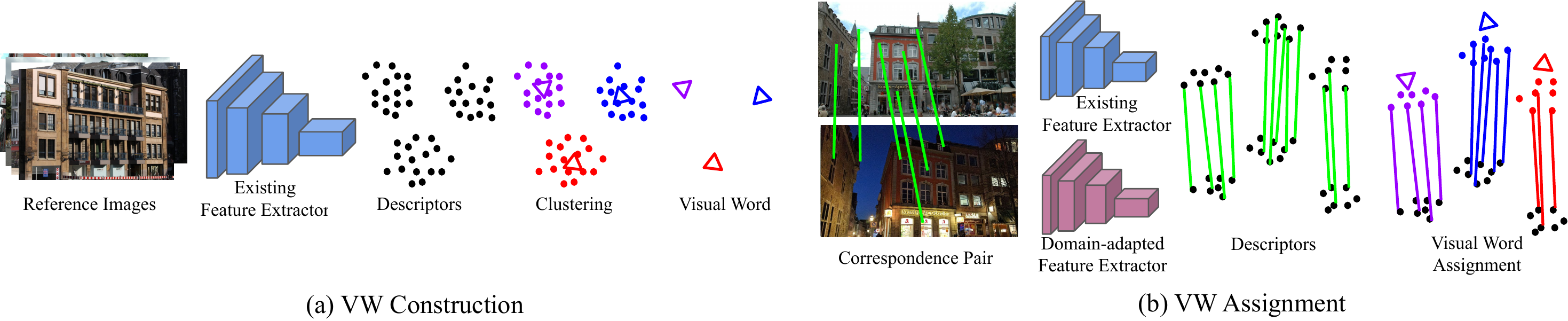}
\vspace{-2em} 
\caption{\small{(a) Visual word construction: The visual words are obtained through K-means clustering on the source descriptors (Sec.~\ref{subsec:vw}).
(b) Visual word assignment: The source and the target descriptors are extracted from a pair of reference and query images. The source descriptors are assigned to the closest visual words. The visual word assignment for a target descriptor is determined by that of its corresponding source descriptor, based on the training feature correspondences (Sec.~\ref{subsec:vwcoral}-\ref{subsec:cdsos}).
}}
\vspace{-1.5em}
\label{fig:VW}
\end{figure*}
\vspace{-0.5em}
\subsection{Overview}\label{sec:overview}
The overview of our approach is shown in Fig.~\ref{fig:overview}. We target the scenario where we are given a 3D point cloud generated from the set of reference images from a known source domain (e.g. daytime) and a few training images from unknown target domain (e.g. nighttime). The point cloud is built from the existing local feature representation and is fixed. The reason why it is kept fixed is because re-building the point cloud from new features takes a significant amount of time and is not feasible in real-world scenarios. When a few training images under an unknown condition are given, we register them to the 3D point cloud to generate training feature correspondences, which can be further refined when the ground-truth pose is given. For reliability of the training data, we only use images that are registered with inliers greater or equal to $15$.
We then use these correspondences to adapt the existing representation to the unknown target domain. Finally, the query images can be registered with better accuracy using the domain adapted features with only a small number of training samples.

\vspace{-1em}
\subsection{Training Objective}
This section describes how we adapt an existing local feature extraction network to a target domain with only few training examples. While our method is applicable to any CNN-based feature extractor, we use the state-of-the-art D2-Net~\cite{dusmanu2019d2} to assess our method in the paper. 
\vspace{-1em}
\subsubsection{Correspondence Loss}\label{subsec:fine_tuning}
As discussed in the overview (Sec.~\ref{sec:overview}), we generate the training correspondence pairs from a small number of training images by registering them to the 3D point cloud. We use these training correspondence pairs to minimize the distance between the source and the corresponding target descriptors in the feature space (positive pairs) and do the opposite for irrelevant pairs (negative pairs). More details on generating the positive and the negative pairs can be found in Sec. \ref{subsec:training_correspondence}.
Specifically, we use the the detection-score-weighted triplet margin ranking loss proposed for D2-Net~\cite{dusmanu2019d2} 
as our correspondence loss:

\begin{equation}
    \mathcal{L}_{\text{Corres}} = \sum_{i=1}^N \frac{s^S_i s^T_{i+}}{\sum_{j=1}^N s^S_{j} s^T_{j+}} \text{max}(0, \|\bm{x}^S_{i} - \bm{x}^T_{i-}\|_2^2- \|\bm{x}^S_{i}-\bm{x}^T_{i+}\|_2^2 + m) \ ,
    \label{eq:loss}
\end{equation}
where $\bm{x}^T_{i+}$ and $\bm{x}^T_{i-}$ are the positive and the negative target descriptors that correspond to the source descriptor $\bm{x}^S_{i}$, respectively. $s_i^S$ and $s_{i+}^T$ are the soft keypoint detection scores for the positive pair, $\bm{x}^S_{i}$ and $\bm{x}^T_{i+}$, $N$ is the number of triplets, and $m$ is the margin.
One baseline that we compare against -- the vanilla fine-tuning -- uses the correspondence loss only.

\vspace{-1.5em}
\subsubsection{Per Visual Word Correlation Alignment Loss (VW-CORAL)}\label{subsec:vwcoral}
However, the fine-tuning alone only leads to over-fitting when trained with very few examples, as we demonstrate in the experimental results (Table~\ref{tab:aachen}-\ref{tab:robotcar_results}).  
To overcome this problem, we propose two additional loss terms to regularize the training of target descriptors.
First, we adopt the CORAL loss~\cite{sun2016deep} to minimize the difference in second-order statistics between the source and the target domain.
Unlike the original CORAL loss, rather than blindly aligning the distributions over all descriptors, we align distributions per coarse visual word (Fig.~\ref{fig:loss}(a)):
      \begin{equation}
      \begin{aligned}
      {\mathcal{L}_{\text{VW-CORAL}}}= \frac{1}{K}\sum_{k=1}^{K}{\frac{1}{4d^2}}{\| \bm{C}_{k}^{S} - \bm{C}_{k}^{T} \|}^2_F ,
      \end{aligned}
      \label{eq:coral}
      \end{equation}
where $\bm{C}_{k}^S$ and $\bm{C}_k^{T}$ are the covariance matrix of the source and the target descriptors that belong to $k$-th visual word, ${\|\cdot\|}_F$ is the Frobenius norm and $d$ is the descriptor dimension. 
For each training correspondence pair, the visual word assignment of the source descriptor determines that of its corresponding target descriptor (Fig.~\ref{fig:VW}(b)). Thus, our loss aligns the distribution of the source and target descriptors that \textit{ought to} belong to the same visual word.

\vspace{-1.5em}
\subsubsection{Cross-Domain Second Order Similarity Loss (CD-SOS)}\label{subsec:cdsos}
We can further preserve the structure of the source domain in the target domain by considering the pairwise relationship between the descriptors in each domain. 
To this end, we introduce the cross-domain second-order similarity (CD-SOS) loss, inspired by the SOS regularizer~\cite{tian2019sosnet}.
The original SOS regularizer was proposed to aid metric learning by making distance between a descriptor pair to be similar to the distance between their positives. 
Similarly, our CD-SOS loss enforces the pairwise distances between training examples $\bm{x}^S_i$ in the source domain, to be similar to the distances between the corresponding examples $\bm{x}^T_{i+}$'s in the target domain (Fig.~\ref{fig:loss}(b)). 
Similar to VW-CORAL, we apply the CD-SOS loss per coarse visual word basis and observe better performance.
Our CD-SOS loss then becomes: 
\begin{equation}
\label{eq:sod_objective}
\mathcal{L}_{\text{CD-SOS}} = \frac{1}{K}\sum_{k=1}^{K}\frac{1}{N_{k}} \sqrt{\sum_{
j \neq i
}^{N_k}{(\|\bm{x}^S_{i(k)} - \bm{x}^S_{j(k)}\| - \|\bm{x}^T_{i(k)+} - \bm{x}^T_{j(k)+}\|)^2}},
\end{equation}
where $\bm{x}^S_{i(k)}$ and $\bm{x}^S_{j(k)}$ indicates the source descriptors that are assigned to $k$-th visual word and $\bm{x}^T_{i(k)+}$ and $\bm{x}^T_{j(k)+}$ are their corresponding target descriptors. $N_k$ is the number of such pairs.

\vspace{-1em}
\subsubsection{Soft Matching Loss (SoftMatch)}\label{subsec:softmatch}
\begin{figure*}[t]
\centering
\includegraphics[width=\textwidth]{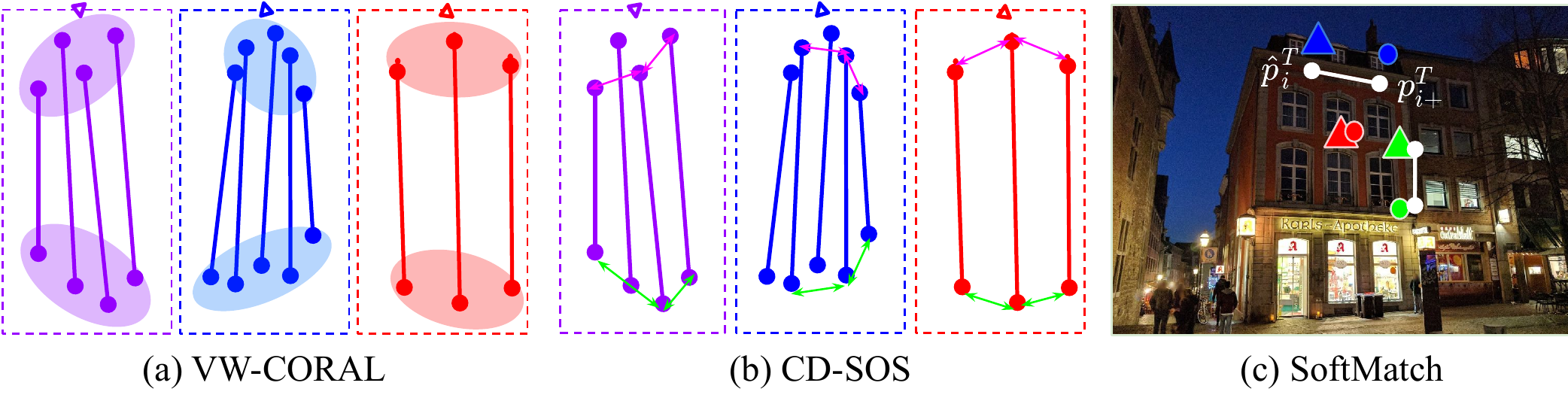}
\vspace{-2em} 
\caption{\small{(a) VW-CORAL loss aligns the distribution of the source and target descriptors for each VW-based group (Sec~\ref{subsec:vwcoral}).
(b) CD-SOS loss enforces the pairwise distances of the target descriptors (\textcolor{green}{green}) to be similar to those of the source descriptors (\textcolor{magenta}{magenta}) for each VW-based group (Sec~\ref{subsec:cdsos}).
(c) SoftMatch loss penalizes its matched keypoints $\bm{\hat{p}}^T_i$ in the target image $I^T$ that are far from the actual corresponding keypoints $\bm{p}^T_{i+}$ (Sec~\ref{subsec:softmatch}).
More details can be found in the supplementary material.
}}
\vspace{-0.5cm}
\label{fig:loss}
\end{figure*}
We observed that minimizing the correspondence loss does not always translate to achieving higher pose estimation accuracy. Therefore, we propose a loss that explicitly penalizes the feature matches that conflict with the given pose during training. 
To make the matching differentiable, we use the soft-argmax function by Luvizon \etal~\cite{luvizon2019human} to perform matching between the source and the target descriptors.
For a source descriptor $\bm{x}^S_i$ at keypoint $\bm{p}^S_i$, we find the location $\bm{\hat{p}}^T_i$ of its match in the target image as follows:
\begin{equation}
    \bm{\hat{p}}^T_i = \textrm{SoftArgMax2D}(\bm{M}(\bm{x}_i^S)),
\end{equation}
where $\bm{M}(\bm{x}_i^S)$ is the 2D heatmap of the matching scores between a source descriptor $\bm{x}^S_i$ and all detected local features $(\bm{x}^T_j, \bm{p}^T_j)$ in the target image $I^T$.
We then compute the distance between the matched feature keypoint $\bm{\hat{p}}^T_i$ and its actual corresponding point $\bm{p}^T_{i+}$, obtained as a result of reprojecting the 3D reference point to $I^T$ using the generated training pose (or ground-truth pose, if available) (Fig.~\ref{fig:loss}(c)). 
To summarize, our soft matching loss is defined as follows, where $l$ is the diagonal length of the image and $n$ is the number of matches:
\vspace{-0.5em}
\begin{equation}
    \mathcal{L}_{\text{SoftMatch}} = {\frac{1}{n}}\sum_{i = 1}^{n}{{\frac{1}{l}} \| \bm{\hat{p}}^T_i - \bm{p}^T_{i+} \|}_2.
\end{equation}
Our overall loss function then becomes as follows, with $\lambda_i$ denoting the weight  for each term:
\vspace{-0.5em}
\begin{equation}
\mathcal{L} = \lambda_1 \mathcal{L}_{\text{Corres}} + \lambda_2 \mathcal{L}_{\text{VW-CORAL}} + \lambda_3 \mathcal{L}_{\text{CD-SOS}} + \lambda_4 \mathcal{L}_{\text{SoftMatch}}.
\end{equation}

\vspace{-1em}
\subsection{Generating Training Correspondence Pairs}\label{subsec:training_correspondence}
To generate the positive training correspondence pairs for a given target image, we perform image registration against the 3D point cloud that was built from the reference images using COLMAP~\cite{schonberger2016structure}. 
We used SIFT~\cite{sift1999lowe} features for this step in the experiments as it was slightly less noisy compared to the D2-Net~\cite{dusmanu2019d2} correspondences. 
While registration using the existing representation may only provide sparse and noisy correspondence pairs, we empirically show that our framework manages to use these sparse correspondences to improve the localization performance through domain adaptation.
The negative pairs are generated via hard negative mining, following \cite{dusmanu2019d2}, which looks for the descriptor with the highest similarity score to the source descriptor, outside a local neighborhood of its positive pair.
\vspace{-1em}
\subsection{Visual Vocabulary Construction}\label{subsec:vw}
For each dataset, we extract local descriptors from randomly sampled reference images using the existing feature extractor (pre-trained D2-Net~\cite{dusmanu2019d2}), and perform K-means clustering to obtain visual words (Fig.~\ref{fig:VW}(a)). In this paper, we used $1$k images and sampled $3$k descriptors, resulting in $3$M descriptors for clustering. 
We evaluated the localization accuracy on the validation set using the different number of visual words ($k=32, 64, 128$), and empirically determined the number of clusters as $64$. This part is done offline before training. 
\vspace{-1em}
\subsection{Implementation Details}\label{subsec:implementation_details}
For the baseline feature extractor, we use the pre-trained D2-Net~\cite{dusmanu2019d2} provided by the authors.
Similar to how D2-Net was originally trained, we optimize the last layer (\texttt{conv4\_3}) and freeze other layers. 
For all our trained models, including fine-tuning, we used Adam~\cite{kingma2014adam} optimizer with learning rate of $10^{-5}$.
To compensate for the sparsity of the generated ground-truth correspondences, we use a larger batch size of $10$ and use full image resolution, whereas cropped images were used with batch size of 1 in \cite{dusmanu2019d2}. 
We set the margin $m$ for the correspondence loss to be 1. 
The weight for each loss term $\lambda_i$ is determined such that the loss values are in a similar range and they sum up to 1.
Specifically, we compute the means $\mu_i$ and standard deviations $\sigma_i$ of each loss term over the training set and set $\lambda_i$ to be $\frac{1}{4(\mu_i + 3 \sigma_i)}$. 

\vspace{-1em}
\section{Experimental Results}
\subsection{Evaluation Protocol}\label{subsec:eval_protocol}
\textbf{Datasets.}
For evaluation, we used the publicly available datasets including RobotCar Seasons~\cite{robotcar2017maddern} and Aachen Day-Night~\cite{6dof2018sattler}. 
The Aachen Day-Night~\cite{6dof2018sattler} dataset consists of ${\sim}4$k reference images and $922$ query images, where $98$ images are captured in night, and the rest are daytime images.
Because the Aachen Day-Night~\cite{6dof2018sattler} dataset does not provide data splits for training and testing, we randomly divide the nighttime query images into disjoint sets and obtain $20$ training, $16$ validation, and $62$ test images. 
No pose is given for the query images.
Recently, Zhang~\etal~\cite{zhang2020reference} have updated the dataset with more accurate ground-truth poses. In this paper, the updated ground-truth poses are used for evaluation.

RobotCar Seasons~\cite{robotcar2017maddern} provides ${\sim}32$k images in total, where ${\sim}20$k of them are reference images, and the rest are the query images are taken under nine different conditions. 
The reference images are provided with known poses, which are used for reconstructing the 3D model. 
We use a variant of the dataset called the CSC (Cross-Seasons Correspondence) RobotCar Seasons dataset~\cite{Larsson_2019_CVPR} that provides ${\sim}5k$ training images with ground-truth poses and the total of $2,072$ test images across the nine conditions. 
Because we consider few-shot domain adaptation, we only use a small subset (${\sim}35$ images per condition) of available training images, similar to the number of training examples used for Aachen Day-Night~\cite{6dof2018sattler}.

\noindent
\textbf{Evaluation Metric.}
Following the standard protocol for above datasets~\cite{robotcar2017maddern,aachen2012sattler,cmu2011badino}, we evaluate the performance by measuring the pose error: the difference between the ground-truth pose ($\bm{c}, \bm{\mathtt{R}}$) and the estimated pose ($\bm{\hat{c}}, \bm{\hat{\mathtt{R}}}$), where pose consists of camera position $\bm{c}$ and camera rotation matrix $\bm{\mathtt{R}}$.
The pose error consists of the position error $\epsilon_t=\lVert \bm{c} - \bm{\hat{c}} \rVert_2$ and orientation error $\epsilon_r=\arccos(\frac{1}{2}(\text{trace}(\bm{\mathtt{R}}^{-1}\bm{\hat{\mathtt{R}}}) - 1))$. $\epsilon_t$ measures the Euclidean distance between the ground-truth and estimated positions, while $\epsilon_r$ measures the the minimum angle needed to align the axes of both rotations~\cite{hartley2013rotation}.
Finally, the pose accuracy is measured by the proportion of query images that are correctly localized within pose error thresholds of (1) $0.25$ $m$ and $2$ deg, (2) $0.5$ $m$ and $5$ deg, and (3) $5$ $m$ and $10$ deg.

\vspace{-1em}
\subsection{Quantitative Results}\label{sec:quan_result}
\begin{table}[t]
\begin{minipage}[t]{\dimexpr.44\textwidth}
\caption{Recall on the subset under night condition from Aachen Day-Night~\cite{aachen2012sattler}
}
\centering
\scalebox{1}{
\begin{tabular}{l||c}
&night\\ \cline{2-2}
Method &{\small \begin{tabular}[c]{@{}c@{}}.25/.50/5.0 (m)\\ 2/5/10 (deg)\end{tabular}}\\
\hline
{\begin{tabular}[l]{@{}l@{}}D2-Net \cite{dusmanu2019d2} + \\ ToDayGAN~\cite{nighttoday2019anoosheh}\end{tabular}} & 38.7 / 62.9 / 79.0 \\ \hline
D2-Net~\cite{dusmanu2019d2} & \textbf{\textcolor{blue}{67.8}} / \textbf{\textcolor{blue}{80.6}} / \textbf{\textcolor{blue}{98.3}} \\ \hline
fine-tune &\textbf{\textcolor{blue}{67.8}} / \textbf{\textcolor{blue}{80.6}} / 97.7 \\ \hline
\textbf{Ours} & \textbf{\textcolor{red}{71.0}} / \textbf{\textcolor{red}{83.9}} / \textbf{\textcolor{red}{100}}\\ \hline
\end{tabular}
}
\label{tab:aachen}
\end{minipage}
\hfill
\begin{minipage}[t]{\dimexpr.52\textwidth}
\centering
\caption{Overall performance on day and night conditions on CSC RobotCar Seasons~\cite{Larsson_2019_CVPR}. The \textcolor{red}{first} and \textcolor{blue}{second} best results are highlighted. 
}
\setlength\tabcolsep{1.1pt}
\scalebox{1}{
\begin{tabular}{l||c|c}
& all-day & all-night \\ \cline{2-3}
Method & {\small \begin{tabular}[c]{@{}c@{}}.25/.50/5.0 (m)\\ 2/5/10 (deg)\end{tabular}} & {\small \begin{tabular}[c]{@{}c@{}}.25/.50/5.0 (m)\\ 2/5/10 (deg)\end{tabular}}\\
\hline
FGSN~\cite{finegrained2019larsson} & \textcolor{red}{\textbf{61.6}} / \textcolor{red}{\textbf{93.5}} / \textcolor{red}{\textbf{99.7}} & 11.0 / 28.4 / 45.2 \\ \hline
D2-Net \cite{dusmanu2019d2} & 59.3 / 89.5 / \textcolor{blue}{\textbf{98.5}} & \textcolor{blue}{\textbf{20.0}} / \textcolor{blue}{\textbf{39.8}} / 49.4 \\ \hline
fine-tune & 59.5 / 89.5 / 98.3 & 19.2 / 39.6 / \textcolor{blue}{\textbf{50.9}} \\ \hline
\textbf{Ours} & \textcolor{blue}{\textbf{61.0}} / \textcolor{blue}{\textbf{89.6}} / \textcolor{blue}{\textbf{98.5}} & \textcolor{red}{\textbf{20.9}} / \textcolor{red}{\textbf{42.6}} / \textcolor{red}{\textbf{52.1}}\\ \hline
\end{tabular}
}
\label{tab:robotcar_2}
\end{minipage}
\hfill
\vspace{-1.5em}
\end{table}
In Table~\ref{tab:aachen}, we compare the localization recall of our method with the baselines on Aachen Day-Night~\cite{6dof2018sattler}. The baselines include fine-tuning which uses the correspondence loss (Eq. \ref{eq:loss}) only, the pre-trained D2-Net~\cite{dusmanu2019d2}, and D2-Net on the night-to-day converted images using ToDayGAN~\cite{nighttoday2019anoosheh}.
Our method significantly outperforms the baselines while only using an average of 20 training images, whereas fine-tuning failed to show any improvement. 

A similar trend is observed for the CSC RobotCar Seasons~\cite{Larsson_2019_CVPR} dataset in Table~\ref{tab:robotcar_2}, where our method consistently outperforms the pretrained D2-Net and fine-tuning.
For this dataset, we also compare our results with the fine-graned segmentation network (FGSN)~\cite{finegrained2019larsson} with semantic match consistency (SSMC)~\cite{toft2018semantic}.
Although we use much smaller portion (${\sim}6.7\%$) of the available training images compared to FGSN~\cite{finegrained2019larsson}, our method yields much better recall on nighttime conditions, while achieving competitive performance on daytime conditions.

In Table~\ref{tab:robotcar_results}, we compare performance of the proposed method and fine-tuning as we change the number of training images, for each weather condition in the CSC RobotCar Seasons dataset~\cite{Larsson_2019_CVPR}.
It shows that fine-tuning often performs worse than the pre-trained model when less training images are available. 
On the other hand, the proposed method consistently achieves better or comparable accuracy.
It can be also seen that our method performs well for nighttime and winter conditions, where the domain gap is most significant.

We continue to compare the proposed method with ToDayGAN~\cite{nighttoday2019anoosheh} on CSC RobotCar Seasons~\cite{Larsson_2019_CVPR} in Table~\ref{tab:todaygan}.  ToDayGAN~\cite{nighttoday2019anoosheh} also aims to reduce the domain gap by translating nighttime into daytime images, thereby greatly improving the image retrieval performance.
However, the translated images contain artifacts that deteriorate the local features, which leads to a decrease in the registration performance.
The proposed method, on the other hand, focuses on adapting the local features to the target domain, without the direct modification of images. 
While ToDayGAN~\cite{nighttoday2019anoosheh} is trained on ${\sim}7k$ images, including $868$ nighttime images from RobotCar Seasons, it is outperformed by our method that uses very few training images.

\setlength{\tabcolsep}{3pt}
\begin{table*}[t]
\begin{center}
\caption{
Recall for each condition on CSC RobotCar Seasons dataset~\cite{Larsson_2019_CVPR}. 
The second column lists the average number of training images $I^T$ of target domain and that of image pairs $(I^S, I^T)$ across the different conditions. \textcolor{red}{first} and \textcolor{blue}{second} best results are highlighted.  
}
\vspace{-0.5em}
\resizebox{\columnwidth}{!}{\begin{tabular}{c|c||c|c|c|c|c}
 & train & {night}&{night-rain}& {dawn}& {dusk} & {OC-summer}\\ \cline{3-7}

Method &{\begin{tabular}[c]{@{}c@{}} avg \#img / \\ \#pair \end{tabular}}& {\begin{tabular}[c]{@{}c@{}}.25/.50/5.0 \small{(m)}\\ 2/5/10 \small{(deg)}\end{tabular}} & {\begin{tabular}[c]{@{}c@{}}.25/.50/5.0 \small{(m)}\\ 2/5/10 \small{(deg)}\end{tabular}} & {\begin{tabular}[c]{@{}c@{}}.25/.50/5.0 \small{(m)}\\ 2/5/10 \small{(deg)} \end{tabular}} & {\begin{tabular}[c]{@{}c@{}}.25/.50/5.0 \small{(m)}\\ 2/5/10 \small{(deg)} \end{tabular}} & {\begin{tabular}[c]{@{}c@{}}.25/.50/5.0 \small{(m)}\\ 2/5/10 \small{(deg)} \end{tabular}} \\ \hline

{D2-Net \cite{dusmanu2019d2}}& - & 21.9 / 42.6 / 51.7 & 17.9 / 36.6 / 46.8 &  55.1 / 90.7 / 99.3 & 76.4/ 94.3/ 100 &  38.3 / 85.5 / 94.4\\ \hline

fine-tune &{\begin{tabular}[c]{@{}c@{}} 13 / \end{tabular}} & 21.1 / 42.3 / 53.2 & 17.1 / 35.4 / 45.5 &  55.1 / 90.7 / 99.3 & 74.5 / 94.4 / 100  & 37.0 / 84.2 / 95.7\\ 

\textbf{Ours} & {\begin{tabular}[c]{@{}c@{}} 305 \end{tabular}} & \textbf{\textcolor{red}{24.4}} / \textbf{\textcolor{blue}{44.8}} / 53.2 & \textbf{\textcolor{blue}{19.1}} / \textbf{\textcolor{red}{40.6}} / \textbf{\textcolor{blue}{47.2}} & 55.1 / 89.4 / \textbf{\textcolor{red}{100}} & \textbf{\textcolor{blue}{77.9}} / 95.9 / 100 & \textbf{\textcolor{blue}{39.5}} / 85.5 / 95.7 \\ \hline

fine-tune &{\begin{tabular}[c]{@{}c@{}} 20 / \end{tabular}} & 22.6 / 42.3 / 52.8 & 17.1 / 37.4 / \textbf{\textcolor{blue}{47.2}} &  55.1 / 90.7 / 99.3 & \textbf{\textcolor{blue}{77.9}} / 94.4 / 100  &  37.0 / 85.5 / 95.7\\

\textbf{Ours} &{\begin{tabular}[c]{@{}c@{}} 610 \end{tabular}} & \textbf{\textcolor{blue}{23.7}} / 44.1 / \textbf{\textcolor{blue}{53.9}} & \textbf{\textcolor{red}{20.3}} / \textbf{\textcolor{blue}{39.8}} / 46.7 &  \textbf{\textcolor{red}{56.4}} / 90.7 / \textbf{\textcolor{red}{100}} & \textbf{\textcolor{red}{79.4}} / 95.9 / 100 &  \textbf{\textcolor{blue}{39.5}} / 85.5 / 95.7\\ \hline

fine-tune &{\begin{tabular}[c]{@{}c@{}} 35 / \end{tabular}} & 21.5 / 41.9 / 52.8 & 16.7 / 37.0 / \textbf{\textcolor{red}{48.8}} &  55.1 / 90.7 / 99.3 & 76.4 / 95.9 / 100  & 38.3 / 85.5 / 94.4\\ 

\textbf{Ours} &{\begin{tabular}[c]{@{}c@{}} 1220 \end{tabular}} & 22.5 / \textbf{\textcolor{red}{45.2}} / \textbf{\textcolor{red}{55.0}} & \textbf{\textcolor{blue}{19.1}} / \textbf{\textcolor{blue}{39.8}} / \textbf{\textcolor{red}{48.8}} & \textbf{\textcolor{red}{56.4}} / 90.7 / 99.3  & \textbf{\textcolor{red}{79.4}} / 95.9 / 100 &  \textbf{\textcolor{red}{40.8}} / \textbf{\textcolor{red}{86.7}} / 95.7 \\ 
\hline
\cline{1-7}

\end{tabular}}
\resizebox{0.8\columnwidth}{!}{\begin{tabular}{c|c||c|c|c|c}
 & train & {OC-winter}&{snow}& {rain}& {sun}\\ \cline{3-6}

Method &{\begin{tabular}[c]{@{}c@{}} avg \#img /\\ \#pair \end{tabular}}& {\begin{tabular}[c]{@{}c@{}}.25/.50/5.0 \small{(m)} \\ 2/5/10 \small{(deg)} \end{tabular}} & {\begin{tabular}[c]{@{}c@{}}.25/.50/5.0 \small{(m)}\\ 2/5/10 \small{(deg)} \end{tabular}} & {\begin{tabular}[c]{@{}c@{}}.25/.50/5.0 \small{(m)}\\ 2/5/10 \small{(deg)} \end{tabular}} & {\begin{tabular}[c]{@{}c@{}}.25/.50/5.0 \small{(m)} \\ 2/5/10 \small{(deg)} \end{tabular}} \\ \hline

{D2-Net \cite{dusmanu2019d2}}& - & 54.4 / 92.3 / 100 & 62.0 / 91.8 / 99.2 &  \textbf{\textcolor{red}{79.2}} / 94.5 / 100 & 53.1 / 77.8 / 95.3\\ \hline

fine-tune &{\begin{tabular}[c]{@{}c@{}} 13 / \end{tabular}} & 54.4 / 92.3 / 100 & 63.2 / 91.8 / 99.2 & 76.5 / 94.5 / 100  & 51.8 / 76.6 / 96.3\\ 

\textbf{Ours} & {\begin{tabular}[c]{@{}c@{}} 305 \end{tabular}} & \textbf{\textcolor{blue}{56.1}} / 92.3 / 100 & 63.2 / \textbf{\textcolor{red}{93.0}} / 99.2 & \textbf{\textcolor{red}{79.2}} / 94.5 / 100 & \textbf{\textcolor{blue}{54.3}} / \textbf{\textcolor{red}{79.0}} / 96.3\\ \hline

fine-tune &{\begin{tabular}[c]{@{}c@{}} 20 / \end{tabular}} & 54.4 / 92.3 / 100 & 63.2 / 91.8 / 99.2 & 76.5 / 94.5 / 100  & 51.8 / 76.6 / 96.3\\

\textbf{Ours} &{\begin{tabular}[c]{@{}c@{}} 610 \end{tabular}} &  \textbf{\textcolor{red}{57.7}} / 90.7 / 100 & \textbf{\textcolor{red}{64.5}} / 91.8 / 99.2 & \textbf{\textcolor{blue}{77.9}} / 94.5 / 100 & \textbf{\textcolor{red}{55.6}} / \textbf{\textcolor{red}{79.0}} / 96.3\\ \hline

fine-tune &{\begin{tabular}[c]{@{}c@{}} 35 / \end{tabular}} & 54.4 / 92.3 / 100 &  \textbf{\textcolor{red}{64.5}} / 91.8 / 99.2 & \textbf{\textcolor{blue}{77.9}} / 94.5 / 100 & 53.1 / 77.8 / 96.3 \\ 

\textbf{Ours} &{\begin{tabular}[c]{@{}c@{}} 1220 \end{tabular}} & \textbf{\textcolor{red}{57.7}} / 92.3 / 100 & \textbf{\textcolor{red}{64.5}} / 91.8 / 99.2 & \textbf{\textcolor{blue}{77.9}} / 94.5 / 100 & \textbf{\textcolor{blue}{54.3}} / 77.8 / 96.3 \\ 
\hline
\cline{1-6}

\end{tabular}}
\label{tab:robotcar_results}
\vspace{-1em}
\end{center}
\end{table*}

\begin{table*}[htb!]
\vspace{-0.5em}
\caption{Comparison with ToDayGAN~\cite{nighttoday2019anoosheh} on CSC RobotCar Seasons dataset~\cite{Larsson_2019_CVPR}. 
We compare the localization recall with two image retrieval methods: DenseVLAD on original query images, and DenseVLAD on ToDayGAN generated images (ToDayGAN + DenseVLAD).}
\begin{center}
\vspace{-0.6em}
\setlength\tabcolsep{1.1pt}
\small
\resizebox{\columnwidth}{!}{\begin{tabular}{c||c|c||c|c} 
 & \multicolumn{2}{c||}{night}& \multicolumn{2}{c}{night-rain}\\ \cline{1-5}
\diagbox{Registration}{Retrieval}&{\begin{tabular}[c]{@{}c@{}} ToDayGAN + \\ DenseVLAD\end{tabular}} & DenseVLAD & {\begin{tabular}[c]{@{}c@{}} ToDayGAN + \\ DenseVLAD\end{tabular}} &  DenseVLAD \\ \hline
ToDayGAN~\cite{nighttoday2019anoosheh} & 0.7 / 9.1 / 56.8 & - & 3.7 / 16.7 / 52.0 & -\\ \hline
ToDayGAN + D2-Net & 25.9 / 53.2 / \textbf{76.9} & 19.7 / 35.3 / 51.7 & 22.0 / 49.6 / 61.0 & 14.6 / 31.3 / 42.7 \\ \hline
\textbf{Ours} & \textbf{26.6} / \textbf{56.1} / 76.5 & \textbf{22.6 / 45.2 / 55.0} & \textbf{28.0 / 54.5 / 66.7} & \textbf{19.1 / 39.8 / 48.8} \\ \hline
\end{tabular}}
\label{tab:todaygan}
\vspace{-1.8em}
\end{center}
\end{table*}

\vspace{-1em}
\subsection{Qualitative Results}
\begin{figure*}[t]
\centering
\includegraphics[width=0.96\textwidth]{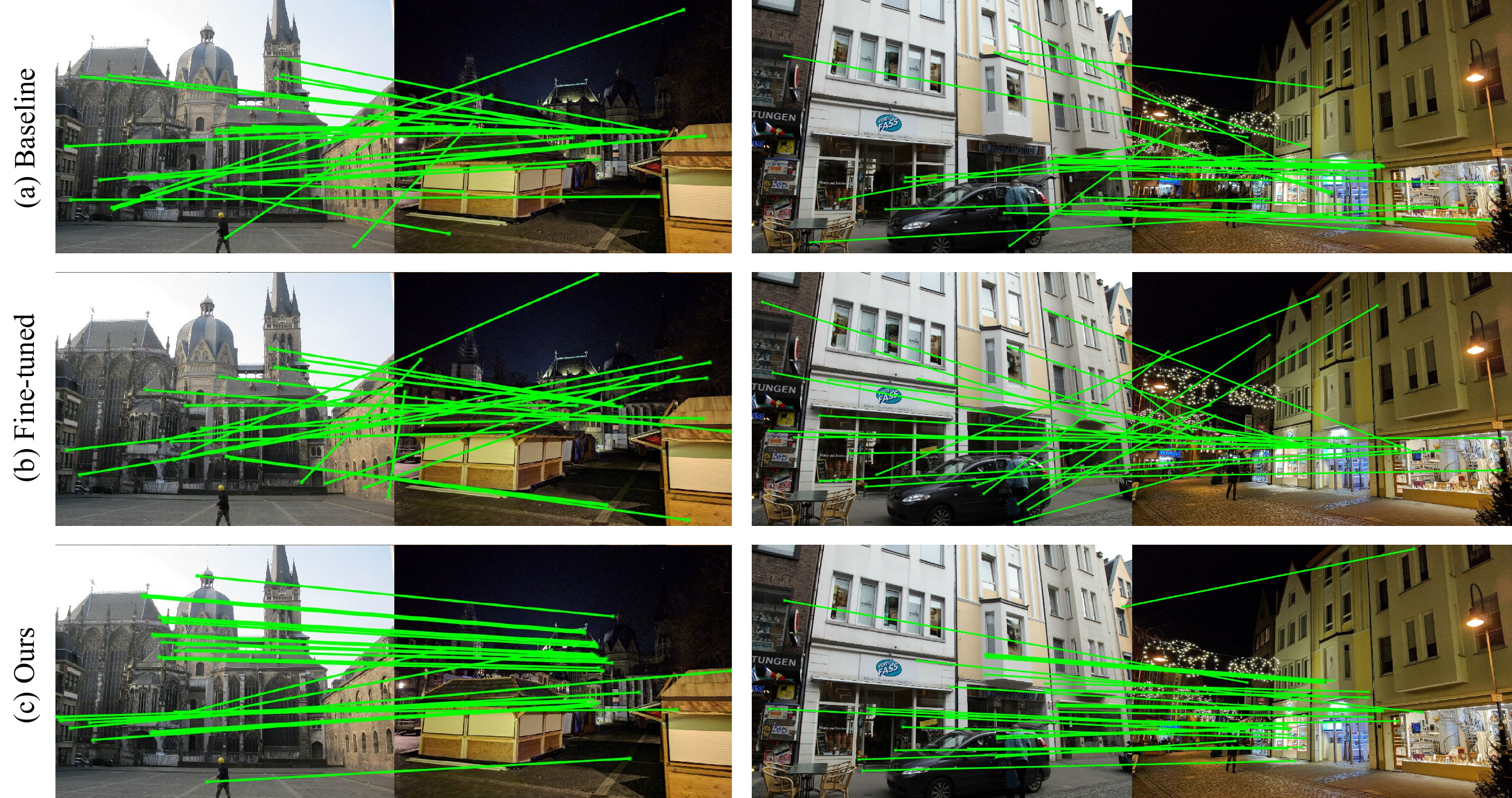}
\vspace{-1em}
\caption{Qualitative results on the Aachen Day Night dataset~\cite{aachen2012sattler} : The inlier matches are visualized for each pair of the retrieved database image (left) and the the query image (right). 
}
\vspace{-0.3cm}
\label{fig:aachen_qual}
\end{figure*}

\begin{figure*}[h!]
\centering
\includegraphics[width=0.97\textwidth]{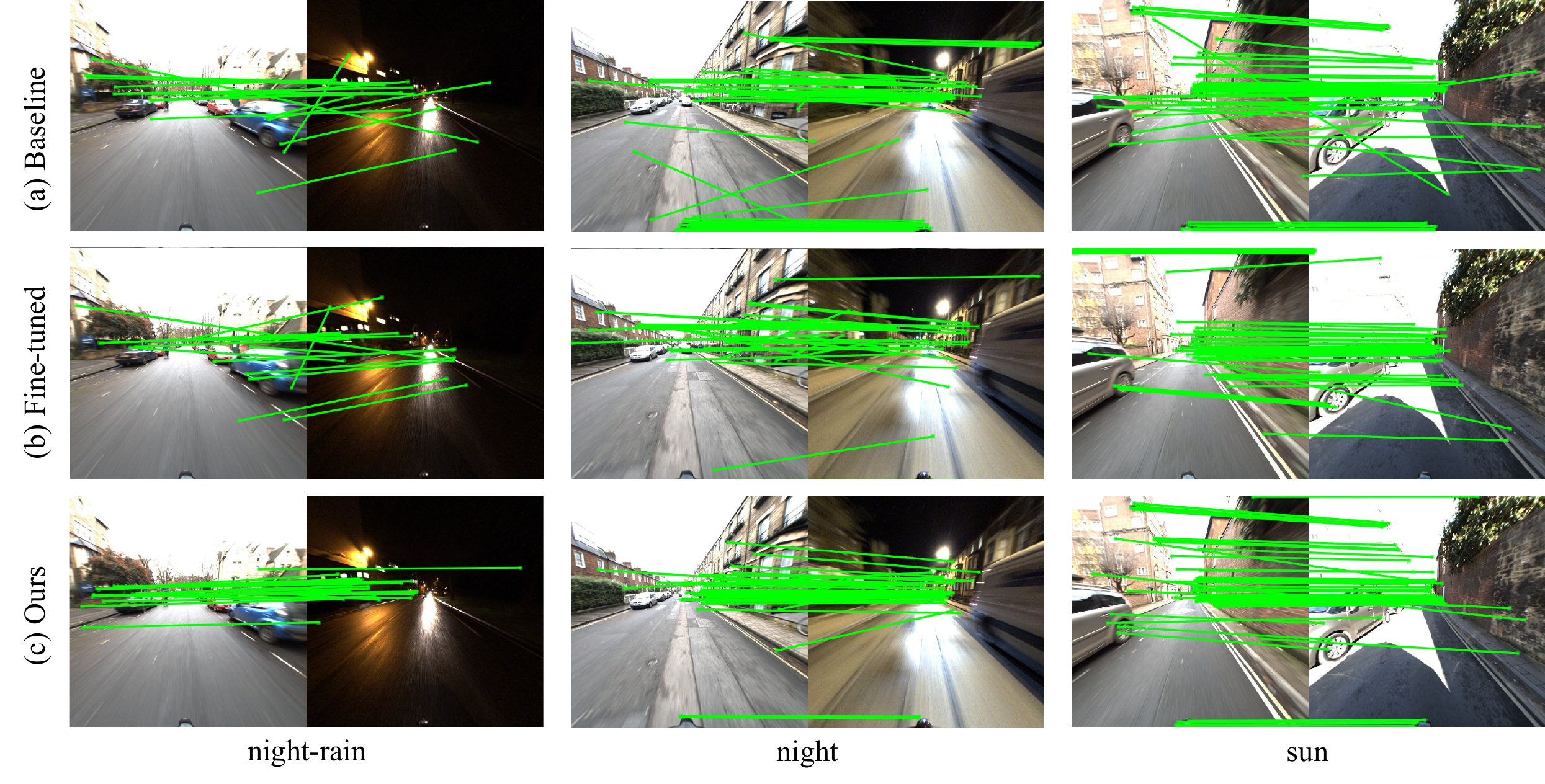}
\vspace{-1em}
\caption{Qualitative results on the RobotCar Seasons Dataset~\cite{robotcar2017maddern}: The inlier matches are visualized for each pair of the retrieved database image (left) and the the query image (right). 
}
\vspace{-0.35cm}
\label{fig:robotcar_qual}
\end{figure*}

The qualitative results on the Aachen Day-Night~\cite{6dof2018sattler} and the CSC RobotCar Seasons~\cite{Larsson_2019_CVPR} datasets are shown in Fig.~\ref{fig:aachen_qual} and Fig.~\ref{fig:robotcar_qual}, respectively. 
In each figure, we visualize the inlier matches between a reference-query pair. We compare our method with the pre-trained D2-Net~\cite{dusmanu2019d2} and fine-tuned models.
It can be seen that the proposed method improves the performance across diverse conditions, while the fine-tuning rarely improves the pre-trained model.


\begin{table*}[h!]
\begin{center}
\caption{Ablation study of different loss term combinations on CSC RobotCar Seasons~\cite{robotcar2017maddern}}
\vspace{0.1em}
\setlength\tabcolsep{1.1pt}
\small
\begin{tabular}{l||c|c|c} 
& all day & all night & mean \\ \hline
$\mathcal{L}_{\text{Corres}}$ (fine-tune) & 59.4 / 89.5 / 98.3 & 19.2 / 39.6/ 50.9 & 50.9 / 78.6 / 87.9  \\ \hline
$\mathcal{L}_{\text{Corres}} + \mathcal{L}_{\text{VW-CORAL}}$ & \textbf{61.0} / 89.5 / 98.5 & 20.6 / 41.1 / 50.0 & 52.4 / 78.9 / 87.8 \\ \hline
$\mathcal{L}_{\text{Corres}} + \mathcal{L}_{\text{CD-SOS}}$ & 60.8 / \textbf{89.8} / \textbf{98.7} & \textbf{21.2} / 41.1 / 50.6 & 52.4 / 79.2 / 88.0 \\ \hline
$\mathcal{L}_{\text{Corres}} + \mathcal{L}_{\text{SoftMatch}}$ & 60.5 / \textbf{89.8} / \textbf{98.7} & 19.8 / 43.2/ 50.3 & 51.8 / \textbf{79.7} / 87.9 \\ \hline
$\mathcal{L}_{\text{Corres}} + \mathcal{L}_{\text{VW-CORAL}} + \mathcal{L}_{\text{CD-SOS}}$ & 60.7 / 89.6 / \textbf{98.7} & 20.7 / \textbf{43.4} / \textbf{52.1} & 52.2 / 79.6 / \textbf{88.3} \\ \hline
{\begin{tabular}[l]{@{}l@{}} $\mathcal{L}_{\text{Corres}} + \mathcal{L}_{\text{VW-CORAL}} + $ \\ $\mathcal{L}_{\text{CD-SOS}} + \mathcal{L}_{\text{SoftMatch}}$ \textbf{(Ours)} \end{tabular}} & \textbf{61.0} / 89.6 / 98.5 & 20.9 / 42.6 / \textbf{52.1} & \textbf{52.5} / 79.4 / \textbf{88.3} \\ \hline
\end{tabular}
\label{tab:ablation}
\end{center}
\vspace{-2em}
\end{table*}

\subsection{Ablation Study}
We evaluate the effectiveness of each proposed loss term in Table~\ref{tab:ablation}, by applying them individually in addition to the correspondence loss in our proposed few-shot domain adaptation framework.
Each proposed loss term provides significant performance improvement over the fine-tuning only.
This indicates the effectiveness of regularization achieved by VW-CORAL and CD-SOS in adapting descriptors to the target domain, given only few examples. 
The performance improvement by SoftMatch also illustrates that it provides stronger supervision for pose estimation, compared to the correspondence loss.
The combination of all loss terms, and the combination of VW-CORAL and CD-SOS provide the best overall performance. 

\section{Conclusion}
\vspace{-0.5em}
In this paper, we investigate the domain gap between the training and the test images for visual localization. We propose a novel few-shot domain adaptation framework for learned local features to address this issue. Experimental results demonstrate that the proposed method works well on challenging scenarios where the training and test data have divergent modality. As the data in real world would always possess a distribution different from the training set, this work presents a extensible and practical solution to the visual localization problem. Future directions include online domain adaptation with streaming data.
\vspace{-1em}
\paragraph{Acknowledgements}
\noindent We would like to thank Torsten Sattler for providing CSC RobotCar Seasons dataset~\cite{Larsson_2019_CVPR} and Asha Anoosheh for providing the results for ToDayGAN~\cite{nighttoday2019anoosheh}.
\newpage

\bibliography{camera_ready}
\newpage

\includepdf[pages={-}]{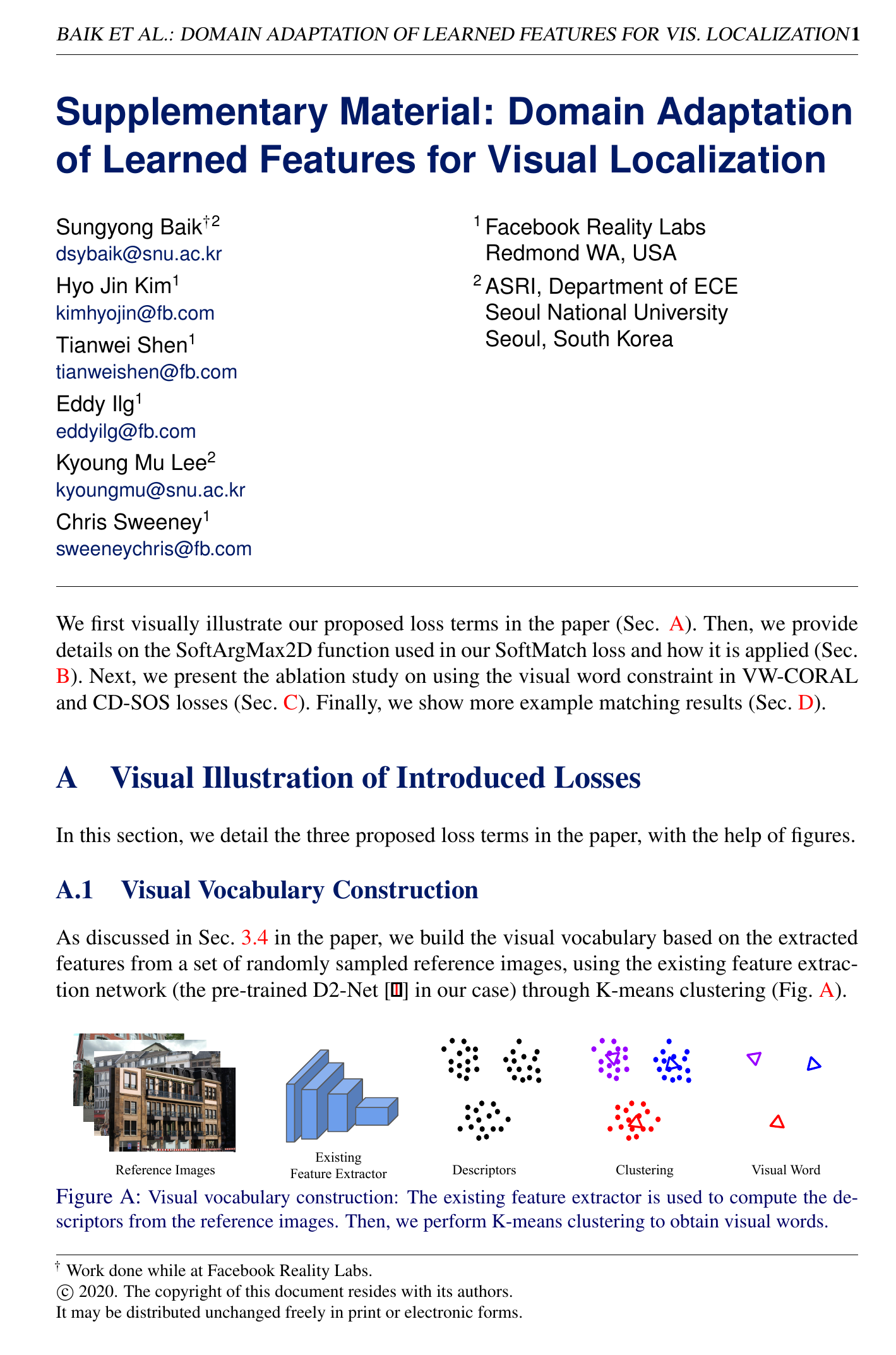}

\end{document}